\PassOptionsToPackage{table}{xcolor}

\documentclass[sigconf]{acmart}
\AtBeginDocument{%
  }
\usepackage{balance}
\usepackage{bigfoot}
\acmConference[Preprint]{Preprint}{2025}{}

\begin{document}

\title{Supporting Preschool Emotional Development with AI-Powered Robots}

\author{Santiago Berrezueta-Guzman}
\email{s.berrezueta@tum.de}
\orcid{0000-0001-5559-2056}
\affiliation{
  \institution{Technical University of Munich}
  \city{Heilbronn}
  \country{Germany}
}

\author{María Dolón-Poza}
\email{maria.dolonp@upm.es}
\orcid{0000-0001-6050-9878}
\affiliation{
  \institution{Universidad Politécnica de Madrid}
  \city{Madrid}
  \country{Spain}}

\author{Stefan Wagner}
\email{stefan.wagner@tum.de}
\orcid{0000-0002-5256-8429}
\affiliation{
  \institution{Technical University of Munich}
  \city{Heilbronn}
  \country{Germany}}

\renewcommand{\shortauthors}{Berrezueta-Guzman et al.}
\renewcommand\footnotetextcopyrightpermission[1]{}

\begin{abstract}

This study evaluates the integration of AI-powered robots in early childhood education, focusing on their impact on emotional self-regulation, engagement, and collaborative skills among preschool children. A ten-week experimental design involving two groups of children assessed the robot's effectiveness through progress assessments, parental surveys, and teacher feedback. Results demonstrated that early exposure to the robot significantly enhanced emotional recognition, while sustained interaction further improved collaborative problem-solving and social engagement. Parental and teacher feedback highlighted high acceptance levels, emphasizing the robot's ease of integration and positive influence on classroom dynamics. This research underscores the transformative potential of AI in education, offering scalable and adaptive tools to address diverse learner needs while complementing traditional teaching methods. The findings advocate for the broader adoption of AI-powered interventions, carefully examining equitable access, ethical considerations, and sustainable implementation. This work sets a foundation for exploring long-term impacts and expanding applications of AI in inclusive and impactful educational settings.
  
\end{abstract}

\keywords{Robotics, preschool education, emotional self-regulation, AI in education, human-robot interaction.}

\received{20 March 2025}
\received[accepted]{15 April 2025}

\makeatletter
\renewcommand\@makefnmark{}
\makeatother

\maketitle
\footnotetext{
This is the author's \textbf{preprint version} of a paper accepted at the 24th ACM Interaction \\Design and Children (IDC) Conference, to be held June 23–26, 2025 in Reykjavík, \\Iceland. The final published version will be available via ACM Digital Library.
}

\section{Introduction}\label{I}

Technology integration in education has reshaped traditional learning environments, offering innovative tools to address diverse challenges \cite{hamzah2024advancing}. From interactive whiteboards to adaptive learning platforms, technological advancements have enriched classroom experiences and personalized education, catering to the unique needs of every learner \cite{kharchenko2024analysis}. Among these innovations, robotics and AI-powered tools have emerged as transformative elements, particularly in early childhood education, where interactive and engaging methodologies foster foundational skills \cite{alam2024integrated}.

Emotional self-regulation is a critical developmental milestone for preschool children, forming the foundation for academic and social success. It enables children to manage emotions, respond to social cues, and navigate challenges effectively, fostering their ability to focus, engage, and adapt in classroom settings \cite{pahigiannis2020peer}. Children who struggle with emotional regulation often face difficulties maintaining attention and participating in classroom activities, which can negatively impact their learning outcomes \cite{gleason2022development}.

Traditional methods for teaching emotional self-regulation, such as role-playing, storytelling, and structured routines, are effective but often require consistent teacher supervision. This can be particularly challenging in larger or under-resourced classrooms \cite{widiastuti2016preschoolers}. These limitations highlight the need for scalable, adaptive solutions supporting emotional development in dynamic learning environments. 

Recent advancements in artificial intelligence (AI) and robotics present unique opportunities to support these developmental processes through interactive and adaptive tools \cite{jarvela2023human}. AI-powered robots offer a promising alternative, combining interactive features and adaptive learning capabilities to provide individualized emotional development support while reducing the burden on educators \cite{berrezueta2025enhancing}. These AI-powered robots equipped with emotional recognition capabilities can respond to children’s needs in real-time, providing personalized feedback that fosters self-awareness and emotional intelligence \cite{yee2024socially}. By engaging children in interactive activities designed to mimic real-life social situations, these robots can help build critical emotional skills. Furthermore, their ability to monitor and analyze behavioral data offers educators valuable insights into each child’s progress, enabling more informed teaching strategies \cite{berrezueta2022artificial}. 

Moreover, robots have the potential to complement traditional teaching methods by acting as co-facilitators in the classroom. Unlike static tools, robots can adapt to varying classroom dynamics, providing consistent engagement and individualized support \cite{buchem2024human}. This adaptability makes them especially effective in addressing the diverse needs of preschool children, including those with learning or developmental challenges \cite{berrezueta2021assessment}. As these technologies evolve, they promise to enhance educational outcomes and create inclusive learning environments that cater to children of all abilities.

Despite these advantages, integrating robotics in early education also presents challenges that warrant careful consideration. Technology acceptance, teacher training, ethical implications, and cost-effectiveness must be addressed to ensure sustainable implementation \cite{leenes2017regulatory}. Understanding how children, educators, and parents perceive these tools is crucial for developing impactful and widely accepted solutions. 

This study evaluates the impact of AI-powered robots on emotional self-regulation and engagement in preschool children while also examining the broader implications of their integration into educational settings. By leveraging real-time emotional analytics and interactive features, these robots provide tailored feedback and dynamic interventions that enhance emotional awareness and promote positive classroom behaviors. Through their adaptive capabilities, AI-powered robots complement traditional teaching methods and support holistic child development by fostering key social and emotional skills essential for long-term success.

\section{Related Work}

Recent advancements in the field of artificial intelligence (AI) have led to the emergence of transformative tools in the realm of early childhood education \cite{yi2024key}, \cite{SU2022100049}, \cite{YANG2022100061}. AI-powered learning tools provide personalized and interactive learning experiences, offering adaptive content tailored to the developmental needs of young children. Several studies support the claim that integrating AI in early childhood education, including language learning tools and interactive storytelling platforms, enhances children's learning and development. Specifically, it has been shown to promote the development of skills such as creativity \cite{tseng2021plushpal}, emotional regulation \cite{wei2020development}, literacy \cite{neumann2020social}, computational thinking \cite{vartiainen2020learning}, \cite{druga2021children}, and collaborative work \cite{kewalramani2021}.

A systematic review by J. Yin et al. examined the use of AI-enabled tools in K-12 education, emphasizing their role in transforming traditional teaching methods through adaptive strategies and real-time assessments. For example, using adaptive reading platforms and gamified learning applications is analyzed like tools for dynamically adjusting content based on individual progress \cite{yim2024artificial}.

The study by W. J. Wei et al. developed the Smart Early Childhood Education Service System (SECESS), which incorporates emotion detection, statistical analysis, and training services. This system was applied to 183 children in kindergartens, including 40 emotion detection games and 40 dialogue sandwich (SD) animations. The results of the study demonstrate that there has been an improvement in SD skills in parents and teachers and that there is a positive correlation with children's emotional competence \cite{wei2020development}. These tools help children manage their emotions and provide educators and parents with insights to better support their development.

The use of AI has been demonstrated to positively impact the quality of teaching. A notable example is the implementation of Intelligent Tutoring Systems (ITS). These systems offer personalized support environments for young learners by adapting the device interface to their specific learning needs. This is achieved by simultaneously assessing children's cognitive states and individual learning requirements \cite{chen2022two}. Consequently, such AI systems can adapt to differing learning speeds, thus rendering them valuable in various classroom settings. Tools must be designed inclusively to ensure equitable access for all learners, particularly those with neurodevelopmental challenges. A review of assistive AI tools emphasizes the importance of careful implementation to support diverse learning needs while promoting equity and fostering innovation in education \cite{barua2022artificial}.

Conversely, the application of robotics in preschool education has been demonstrated to engender distinctive advantages in fostering classroom activity engagement, enhancing social interaction and active learning among young children of a similar age group \cite{10.1145/3290605.3300677}, \cite{wainer2014using}. For instance, a study by J. Yin et al. investigated the role of a robotic assistant in teaching English to 83 children for whom Chinese was their native language. The activities were structured to include random tasks focusing on verbal comprehension, gestural communication, and a mixture of both. The study revealed significant improvements in vocabulary acquisition, especially in the tasks focusing on verbal comprehension \cite{yin2024influence}.

Another critical aspect of robotic-assisted learning in early childhood education is its potential to improve peer interactions, especially among children who are naturally more reserved. Shyness is frequently associated with reduced social interactions, negatively impacting language development. In a research study conducted by Tolksdorf et al., the impact of shyness on 28 children aged between 4 and 5 years was examined. The study found that shy children participated less frequently in expressive interactions than their more outgoing peers. However, both groups demonstrated comparable gains in vocabulary \cite{10.3389/frobt.2021.676123}.

The integration of AI techniques with robotics has attracted significant societal interest, particularly in the context of early childhood education. Recent research has extensively explored the critical ways in which  AI-powered robots can promote both social and emotional development through interactive play \cite{kory2017flat}, \cite{10.1145/3173386.3177059}, \cite{van2022social}, \cite{10.3389/frobt.2021.676123}.

There has been an increasing use of AI technology in supporting the emotional self-regulation of preschool children \cite{jarvela2023human}, \cite{bettis2022digital}, \cite{vistorte2024integrating}. The emotional self-regulation of preschool children might be considerably improved through the effective application of AI-powered robots \cite{yee2024socially}. Emotion recognition algorithms embedded in AI systems can identify children's emotional states in real-time and offer tailored responses, such as calming activities or positive reinforcement. Kewalramani et al. conducted an in-depth study exploring using interactive AI robotic toys in early childhood education to develop inquiry skills in children aged 4-5 \cite{kewalramani2021}. The data revealed how children worked creatively together to build a sustainable city for their robot and family, fostering creative, emotional, and collaborative inquiry skills. The study highlights the potential of AI toys to develop children's inquiry skills and suggests training teachers to integrate them into educational settings.

Another study presented PopBots, an innovative early childhood AI platform designed to teach preschool children fundamental AI concepts. The study suggests that early AI education can empower children with the skills to comprehend and navigate the expanding presence of AI in their lives. Children's learning outcomes were evaluated using custom AI assessments involving interactive activities with social robots. The median score in the evaluation was 70\%, with children showing the most substantial understanding of knowledge-based systems. The study also examined how these activities influenced children's perceptions of robots. The younger children saw robots as toys and more intelligent than themselves, while older children viewed them as less intelligent but more human-like \cite{10.1145/3290605.3300677}.

\section{Methodology}\label{M}

The methodology section outlines the design and implementation of the study. It details the robot's architecture and data flow.

\subsection{Robot architecture}

The robotic assistant used in this study is built on an open-source design and open-source code that incorporates advanced features from socially assistive robotics. These include AI-powered capabilities like natural language processing and real-time emotional analytics, making the robot more adaptable and practical for educational settings \cite{berrezueta2024exploring}.

\begin{figure}[H]
	\centering
	\includegraphics[width=\linewidth]{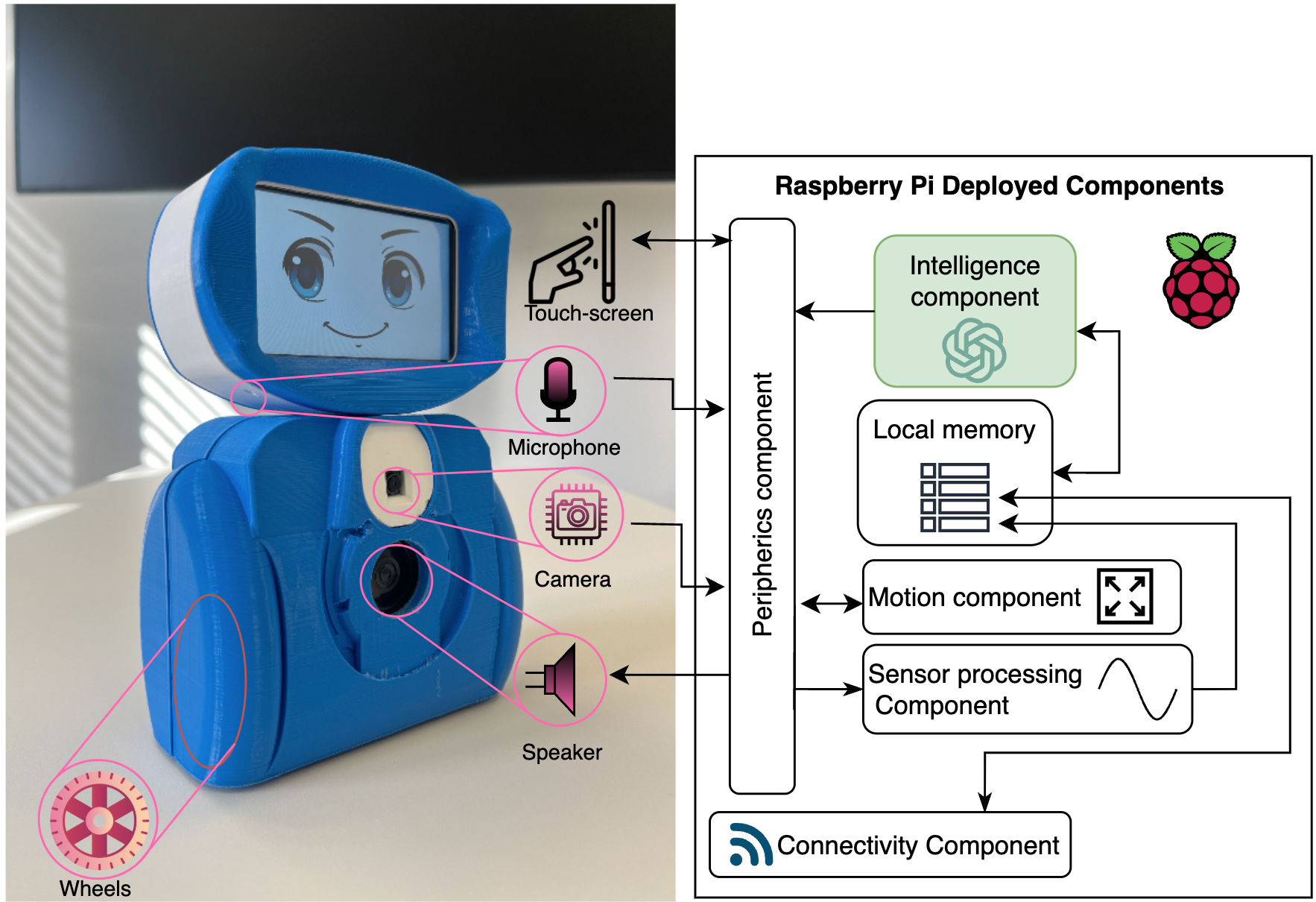}
	\caption{Architecture of the AI-powered robotic assistant, featuring interactive peripherals (left) and a Raspberry Pi-based system (right).}
	\label{robot}
\end{figure}

Figure \ref{robot} illustrates the architecture and key components of the AI-powered robot utilized in the study. On the left, the robot's physical design is depicted, showcasing its interactive features such as a touch-screen display, microphone, camera, speaker, and wheels for mobility. These peripherals enable multi-modal interaction, allowing the robot to engage with preschool children through visual, auditory, and tactile stimuli. On the right, the diagram highlights the internal architecture powered by a Raspberry Pi, which integrates various functional modules. The intelligence component, driven by AI algorithms (GPT-4V(ision)), processes real-time data captured by the camera and microphone for emotional and behavioral analysis. The motion component facilitates mobility and interaction, while the sensor processing unit interprets environmental inputs to adapt the robot’s responses. A local memory module stores activity data and emotional metrics, ensuring adaptive feedback. Finally, the connectivity component supports integration with external systems for data transmission and remote updates, demonstrating the robot’s versatile design tailored for educational settings.

The primary limitations of the Raspberry Pi (low computational power, limited memory, and lack of dedicated GPUs) are addressed by carefully partitioning tasks between the Raspberry Pi and cloud services. Therefore, the robot architecture combines the Raspberry Pi for local pre-processing with GPT-4V(ision), which operates on GPU-supported cloud servers and is accessed via a secure REST API \cite{wu2023early}. This hybrid approach leverages Raspberry Pi's strengths while offloading computationally intensive tasks to the cloud \cite{hurst2024gpt}. 
To manage processed data efficiently, AWS S3 was employed, ensuring seamless integration, scalability, and accessibility for storing emotional metrics and interaction logs when required.

\subsection{Data Flow}

\begin{figure}[H]
	\centering
	\includegraphics[width=\linewidth]{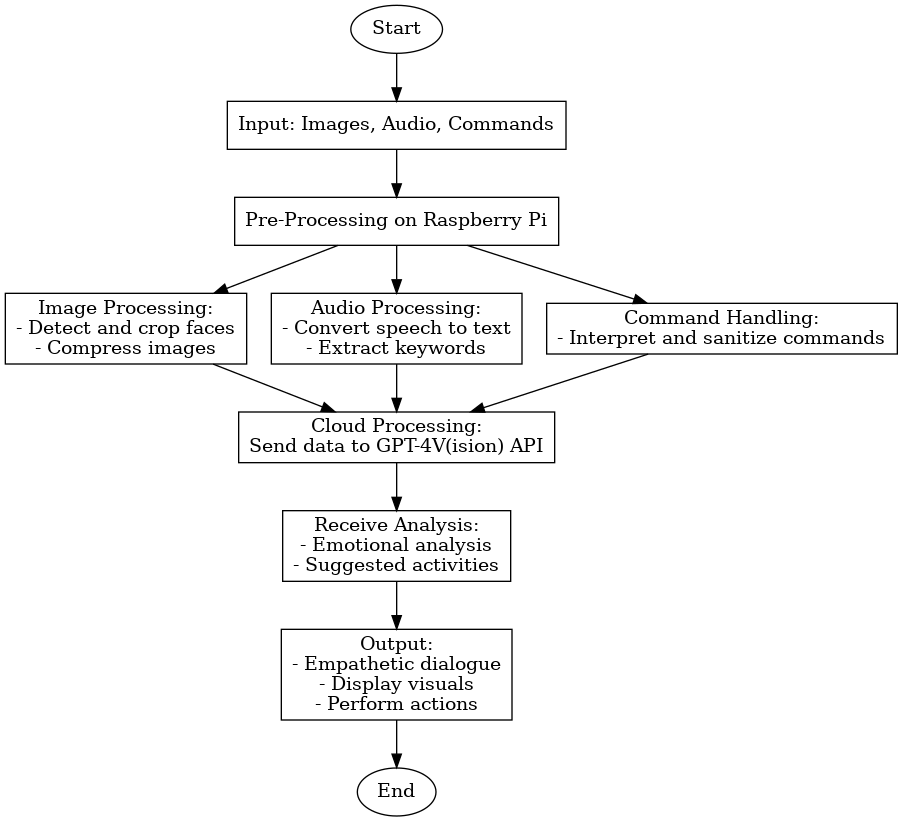}
	\caption{Activity diagram illustrating the data flow process for the AI-powered robot.}
	\label{dataflow}
\end{figure}

Figure \ref{dataflow} illustrates the data flow that begins with input collection, which includes capturing images (children’s facial expressions), audio (speech from children), and commands (buttons on the touchscreen). On the Raspberry Pi, pre-processing steps ensure efficient handling of these inputs. OpenCV detects and crops face for images, and then these images are compressed to reduce data size and allow faster transmission. Audio inputs are converted to text using Whisper API that contains the voice-to-speech API (commonly called speech-to-text API) that works seamlessly with other OpenAI models, including GPT-4V(ision), enabling end-to-end conversational AI solutions with both speech and text inputs \cite{radford2023robust}.

The pre-processed data is sent to the GPT-4V(ision) API via a secure cloud connection for advanced analysis. Images are analyzed for emotion recognition, and text is processed to generate personalized responses or interpret contextual meaning. The API returns outputs such as emotional analysis, suggested activities, or conversational responses, which are then used to drive the robot’s behavior. This includes generating empathetic dialogue through text-to-speech synthesis, displaying relevant information or images on the touchscreen, and performing actions tailored to the interpreted emotional states, such as suggesting calming activities to foster emotional self-regulation.

\section{Experiment Design}

A structured ten-week experimental design was implemented to evaluate the effectiveness of an AI-powered robot in fostering emotional self-regulation and engagement. The study involved multiple stakeholders, including preschool children, teachers, parents, and emotional development specialists, ensuring a comprehensive approach. The methodology incorporated performance assessments, observational studies, and feedback mechanisms to analyze the robot's impact on emotional awareness, classroom participation, and teamwork. The experimental design aimed to provide robust data while maintaining consistency and minimizing external biases across both study groups.

\subsection{Sample Selection}

This study selected two groups of preschool children, each consisting of 20 participants aged 4 years, recruited from different educational institutions. The children came from monolingual households where Spanish was the primary language, providing a uniform linguistic background to eliminate potential confounding factors related to language diversity. 

Gender was recognized as a significant variable, and deliberate efforts were made to ensure a balanced representation of boys and girls in each group. This approach allowed the study to investigate potential gender-based differences in developing emotional self-regulation and engagement outcomes.

\subsection{Description of the experiment}

This experiment assessed the effectiveness of an AI-powered robot in fostering emotional self-regulation and engagement among preschool children. The study also evaluated the acceptance and perceptions of the robot among children, parents, teachers, and school administrators to understand its broader impact on the educational ecosystem.

The experiment lasted ten weeks. Group 1 used the robot for the first five weeks, focusing on activities to improve emotional awareness, classroom participation, and working with peers. Group 1 continued their regular classes for the remaining five weeks, with teachers and parents observing their behavior and progress without further robot interaction in the school and house. 

Group 2 started the experiment with regular preschool activities observed naturally without robot involvement during the first five weeks. From the fifth week onward, they began using the robot, which was integrated into their activities until the end of the study.

Both groups followed a standardized curriculum for emotional recognition, self-regulation strategies, and collaborative tasks. The robot recognized and interpreted children’s emotions by analyzing facial expressions and speech during interactions. It facilitated social scenario simulations by suggesting activities or guiding children through collaborative tasks to enhance emotional self-regulation and teamwork. The robot also provided real-time feedback, dynamically adjusting its responses based on the children’s emotional states or inputs to create a more personalized and engaging experience. Interaction logs, including time spent and emotional metrics, were systematically collected for analysis, and feedback from parents and teachers was gathered to evaluate the robot’s effectiveness and impact on the children’s emotional and social development.

Teachers received comprehensive training on integrating the robot into their lesson plans to ensure robust implementation, including utilizing its real-time emotional analytics and adaptive feedback mechanisms. Emotional development specialists observed sessions biweekly, providing feedback to teachers and fine-tuning activities based on the children’s progress.

The robot’s activities were designed to progressively increase in difficulty and complexity, aligning with the children’s developmental stages. Tasks ranged from identifying and naming emotions to participating in guided group scenarios requiring teamwork and problem-solving. Additionally, the robot recorded emotional metrics, such as changes in facial expressions and verbal interactions, which were analyzed to assess the effectiveness of interventions. Throughout the experiment, the robot remained unchanged regarding hardware and software. This decision was made to ensure consistency and eliminate bias between the groups.

Parents were engaged throughout the study, attending biweekly informational sessions to receive updates on their children’s progress and participating in workshops on reinforcing emotional regulation skills at home. 

The multi-actor approach ensured a holistic evaluation of the robot’s role in promoting emotional self-regulation and classroom engagement while addressing potential challenges and opportunities for future integration. 

\subsection{Research Questions and Hypothesis}

\textbf{Early Emotional Impact.} To highlight the immediate influence of the robot on emotional awareness, engagement, and teamwork, we propose the \textbf{Research Question 1:} How did Group 1, which started using the robot, perform compared to Group 2, which followed traditional methods during the first five weeks? \textbf{Hypothesis 1:} Group 1 will demonstrate higher scores in identifying and naming emotions, suggesting that the robot effectively enhanced early emotional recognition.

\textbf{Adaptation and Progress.} To evaluate how quickly children adapt to the robot and its impact on social interactions, we propose the \textbf{Research Question 2:} How did Group 2, which transitioned to using the robot after the fifth week, compare to Group 1, which had already gained experience with it? \textbf{Hypothesis 2:} Group 2 will show rapid adaptation to the robot, narrowing the progress gap for emotional recognition and social scenario responses. Group 1 is expected to exhibit more consistent engagement and higher accuracy in suggesting appropriate actions during role-play scenarios.

\textbf{Sustained Development.} To assess the long-term effects of the robot on emotional self-regulation and teamwork, we propose the \textbf{Research Question 3:} How did Group 1, which transitioned back to traditional methods, perform in the final assessment compared to Group 2, which continued using the robot? \textbf{Hypothesis 3:} Group 1 will retain many skills acquired during the initial intervention, such as communication and emotional recognition in group tasks. Group 2 will demonstrate superior collaborative skills and sustained engagement performance, emphasizing the long-term benefits of continuous robotic interaction.

\subsection{Progress assessments}

To validate the hypothesis, three progress assessments (PAs) were conducted at weeks 3 (PA1), 7 (PA2), and 10 (PA3). These assessments evaluated the children’s emotional self-regulation, engagement, and teamwork progress. The assessments involved observing the children’s interactions during structured tasks, using specific metrics to measure their ability to recognize emotions, collaborate effectively, and remain engaged in activities.

\textbf{PA1: Emotion Recognition Assessment.} Assessed the children’s ability to identify and name emotions based on visual cues presented by the robot. Points were awarded as follows: 2 points for correctly identifying an emotion, 1 point for partially correct identification (e.g., recognizing general emotional tone but not specific emotion), and one additional point for providing an appropriate verbal explanation or response. Each child was presented with 10 visual cues, resulting in a maximum possible score of 30 points.

\textbf{PA2: Social Interaction Response Assessment.} Evaluated the children’s ability to respond to social scenarios facilitated by the robot—tasks involved recognizing peer emotions during role-play and suggesting appropriate actions. Scoring was based on three metrics: accuracy (up to 10 points, with 1 point per correct identification), engagement level (up to 5 points, based on active participation and verbal contributions), and appropriateness of suggested actions (up to 10 points, with 2 points for each correct or relevant suggestion). The maximum possible score for this assessment was 25 points.

\textbf{PA3: Collaborative Problem-Solving Assessment.} Measured the children’s ability to work collaboratively in guided group tasks. Using scenarios designed by the robot, children were tasked with solving simple problems requiring teamwork. Progress was graded on three criteria: communication (up to 10 points, with 1 point per meaningful contribution), emotional recognition within the group (up to 5 points, with 1 point for each correctly identified group emotion), and problem-solving contribution (up to 10 points, based on the child’s active participation and helpfulness). The maximum possible score for this assessment was 25 points.

These progress assessments comprehensively evaluate the robot’s impact on fostering emotional awareness, engagement, and collaborative skills. Data from the assessments, combined with the emotional metrics recorded by the robot, provide valuable insights into the effectiveness of the interventions and the children’s developmental progress.

\subsection{Parental acceptance evaluation}

Additionally, parents of children in both groups were surveyed to gather their perceptions of the robot’s role in fostering emotional self-regulation and engagement. These surveys were conducted at the end of weeks 5 (S1) and 10 (S2) to capture initial impressions and reflections on the overall experience. The feedback provided valuable insights into how parents viewed the integration of this technology into their children’s education and its perceived impact on their developmental progress.

The survey S1 was conducted after week 5 to gather initial impressions. Some of the questions were:

\begin{enumerate}
    \item How comfortable are you with using a robot in your child’s classroom for emotional development? \\
    (Scale: Very Uncomfortable, Uncomfortable, Neutral, Comfortable, Very Comfortable)
    \item Should we continue using the robot to enhance emotional awareness and engagement in the classroom? \\
    (Yes/No/Unsure)
    \item Have you noticed any changes in your child’s emotional awareness or engagement at home since the introduction of the robot? \\
    (Scale: Strongly Disagree, Disagree, Neutral, Agree, Strongly Agree)
\end{enumerate}

The survey S2 was conducted after week 10 to reflect on the overall experience and long-term impressions. Some of the questions were:

\begin{enumerate}
    \item Did your child enjoy interacting with the robot during their classroom activities? \\
    (Yes/No/Unsure)
    \item How effective do you believe the robot was compared to traditional methods (e.g., teacher-led discussions or role-playing)? \\
    (Scale: Much Less Effective, Less Effective, Same, More Effective, Much More Effective)
    \item Based on your experience, would you recommend the continued use of the robot in early education for emotional development? \\
    (Yes/No/Unsure)
\end{enumerate}

This parental input complemented the quantitative data collected during the progress assessments, offering a broader perspective on the robot’s effectiveness and acceptance. Additionally, parents were encouraged to share open-ended feedback, which provided nuanced insights into how the robot influenced their children’s behavior and emotional growth at home. 

Together, the surveys enriched the overall evaluation, ensuring that both objective progress metrics and subjective parental perceptions were considered in assessing the robot’s impact.

\subsection{Educators acceptance evaluation}

A teacher survey (TS) was conducted with the participating educators to evaluate the AI-powered robot's usability, effectiveness, and acceptance in fostering emotional self-regulation and engagement among preschool children. The survey provided insights into how the robot influenced classroom dynamics and teaching methods, focusing on several key areas to ensure a comprehensive understanding of the robot’s integration into the educational environment.

\textbf{Usability and Ease of Integration:} 
Teachers were asked how seamlessly the robot could be incorporated into their lesson plans and whether they faced any technical challenges. Questions evaluated the user-friendliness of the robot’s interface and its adaptability to existing classroom routines.

\textbf{Perceived Effectiveness:} 
Teachers rated the robot’s effectiveness in enhancing emotional recognition, engagement, and teamwork among students. Specific emphasis was placed on its ability to foster active participation and improve classroom interactions during role-play and collaborative tasks.

\textbf{Impact on Student-Teacher Interaction:}
The survey explored how the robot affected teacher-student interactions, notably whether it facilitated better classroom management or altered the dynamics of teacher-student relationships.

\textbf{Acceptance and Willingness to Continue Usage:}
Teachers provided their overall impressions of the robot, including their comfort with using the technology and their willingness to continue using it in future classroom settings. They also shared whether they would recommend the robot for other educational contexts or subjects.

\textbf{Feedback and Suggestions:}
Open-ended questions encouraged teachers to share their observations about the robot’s strengths and areas for improvement. Teachers were invited to propose enhancements for future robot iterations, ensuring it meets diverse classroom needs more effectively.

\section{Results}\label{R}

The results section provides a detailed evaluation of the impact of the AI-powered robot on preschool children's emotional self-regulation, engagement, and collaborative skills. 

\subsection{Progress Assessments}

The three progress assessments (PA1, PA2, PA3) revealed insights into the intervention's short-term and long-term effects.

\textbf{PA1 - Emotion Recognition Assessment (Week 3):}
At the first progress assessment, Group 1, which started using the robot immediately, achieved an average score of 26/30 points, demonstrating their ability to recognize specific emotions and provide appropriate verbal responses. In contrast, Group 2, following traditional methods, scored an average of 18/30 points, relying more on general emotional tones rather than precise identifications. These results highlight the early advantages of integrating the robot into emotional development activities, validating the \textit{Early emotional impact} hypothesis. 

\textbf{PA2 - Social Interaction Response Assessment (Week 7):}
By the second assessment, Group 2 showed rapid improvement after transitioning to the robot, achieving an average score of 22/25. Their progress reflected significant gains in recognizing peer emotions and suggesting appropriate actions during role-play activities. Meanwhile, Group 1 maintained consistent performance with an average score of 23/25, demonstrating sustained engagement and accuracy in social interactions and validating the \textit{Adaptation and Progress} hypothesis. 

\textbf{PA3 - Collaborative Problem-Solving Assessment (Week 10):}
The final assessment revealed a contrast in long-term outcomes. Group 1, which returned to traditional methods, scored 21/25, retaining many skills acquired during the initial robot interaction but showing a slight decline in problem-solving contributions and emotional recognition within group tasks. Group 2, which continued using the robot, scored 24/25, exhibiting superior communication, teamwork, and sustained engagement, validating the \textit{Sustained Development} hypothesis.

\subsection{Parental Surveys}

Parental surveys conducted at weeks 5 (S1) and 10 (S2) captured evolving perceptions of the robot:

\textbf{Survey S1 Results (Week 5):}
After Week 5 (S1), 78\% of Group 1 parents observed notable improvements in their children’s emotional awareness at home, in contrast to only 40\% of Group 2 parents, reflecting the early impact of the robot. Additionally, 85\% of Group 1 parents expressed comfort with the robot's integration into classroom activities, attributing their approval to the increased engagement and enthusiasm demonstrated by their children. Notably, 90\% of Group 1 parents supported the continued use of the robot in educational settings, emphasizing its perceived effectiveness in fostering emotional development and enhancing classroom interactions.

\textbf{Survey S2 Results (Week 10):}
By the end of Week 10, 93\% of Group 2 parents reported significant improvements in their children’s emotional engagement and social interactions, highlighting the robot's effectiveness during the later phase of the study. Additionally, 88\% of Group 2 parents rated the robot as more effective than traditional methods in fostering emotional development. Across both groups, an overwhelming 95\% of parents recommended integrating the robot into other educational activities, underscoring its potential for broader applications in early education.

\subsection{Teacher Feedback}

The teacher surveys provided valuable insights into the usability, effectiveness, and classroom dynamics influenced by the robot:

\textbf{Usability and Integration:} All teachers found the robot easy to integrate into lesson plans, with minimal technical challenges reported.

\textbf{Effectiveness:} Teachers observed increased participation and focus during robot-facilitated activities, particularly in role-play scenarios.

\textbf{Impact on Classroom Dynamics:} The robot encouraged independent engagement among students, allowing teachers to focus on observing and providing targeted support.

\textbf{Suggestions for Improvement:} Teachers recommended expanding the range of scenarios and enhancing the robot’s adaptability to real-time responses.

\section{Discussion}\label{D}

The findings of this study provide robust evidence supporting the integration of AI-powered robots in early childhood education, particularly in fostering emotional self-regulation, engagement, and collaborative skills. The results demonstrate the robot's capacity to significantly enhance emotional awareness and social interactions through its adaptive, interactive, and engaging methodologies.

\textbf{Immediate Benefits of Robotic Integration:} The early advantages of incorporating the robot were evident in the first progress assessment (PA1), where Group 1, utilizing the robot from the outset, achieved significantly higher scores in emotion recognition than Group 2. This result validates the hypothesis that early exposure to AI-powered tools can effectively accelerate the development of emotional awareness in young children. Parental feedback from Week 5 corroborates this finding, with most reporting noticeable improvements in children’s emotional understanding and engagement at home.

\textbf{Adaptation and Rapid Progress:} By Week 7, Group 2, which transitioned to using the robot after the initial period of traditional methods, demonstrated rapid adaptation and significant gains in social interaction skills. This progress highlights the robot's accessibility and ability to effectively engage children regardless of its introduction. Group 1, meanwhile, maintained consistent performance, underscoring the tool’s potential for sustained engagement. This supports the hypothesis that children can quickly adapt to interactive robotic systems, resulting in measurable improvements in social and emotional learning outcomes.

\textbf{Sustained Development and Long-term Impact:} The final assessment (PA3) revealed notable differences in the long-term effects of continuous versus interrupted robot use. Group 2, which continued interacting with the robot, displayed superior collaboration and sustained engagement compared to Group 1, which reverted to traditional methods. These findings confirm that prolonged exposure to AI-powered interventions yields more profound and lasting benefits, particularly in complex tasks such as collaborative problem-solving.

\textbf{Perceptions and Acceptance Among Stakeholders:} The overwhelmingly positive feedback from parents and teachers underscores the robot’s practical feasibility and acceptance in educational settings. Parents expressed high satisfaction with the robot's impact on their children, with many recommending its integration into broader educational contexts. Teachers highlighted its ease of integration into lesson plans and its effectiveness in enhancing classroom participation, particularly in dynamic group activities. Suggestions for future improvements, such as expanding scenario options and refining real-time adaptability, offer valuable insights for optimizing the tool's design.

\textbf{Challenges and Considerations:} Despite the positive outcomes, the study also highlights critical considerations for sustainable implementation. Ensuring equitable access to such technologies, addressing potential biases in emotion recognition algorithms, and providing adequate teacher training are essential for scaling this approach across diverse educational environments. Additionally, while the robot demonstrated significant benefits, ensuring it complements rather than replaces traditional teaching methods is crucial, preserving educators' role as central learning facilitators.

\textbf{Implications for Future Research and Practice:} This study emphasizes the transformative potential of AI-powered robots in education and sets the foundation for future research exploring broader applications, including interventions for children with developmental challenges or those in underserved communities. Longitudinal studies assessing the sustained impact of such technologies beyond the classroom could further validate their role in holistic child development.

\section{Conclusions}\label{C}

This study highlights the transformative potential of integrating AI-powered robots into early childhood education, particularly in fostering emotional self-regulation, engagement, and collaborative skills. By leveraging real-time emotional analytics and adaptive interaction capabilities, these robots offer innovative solutions to longstanding educational challenges, such as limited teacher resources and diverse learner needs.

The results demonstrate that early exposure to AI-powered tools significantly enhances preschool children's emotional recognition and social interaction skills. Children quickly adapted to the robot, displaying measurable progress in collaborative tasks and problem-solving abilities, even when introduced later in the study. Sustained interaction with the robot further amplified these benefits, underscoring the importance of continuous use to maximize developmental outcomes.

Parental and teacher feedback revealed strong support for the robot's integration, citing increased engagement, enhanced classroom dynamics, and observable improvements in emotional awareness. The findings validate such technologies' practical feasibility and acceptability in educational settings, paving the way for broader adoption.

Despite the promising results, the study also identified key considerations for sustainable implementation, including equitable access, teacher training, and the ethical design of emotion recognition systems. These challenges must be addressed to ensure that AI-powered interventions are inclusive, effective, and aligned with best practices in education.

Looking ahead, this research opens avenues for exploring the long-term impact of AI in education and its applications for diverse populations, including children with developmental challenges. Future studies should evaluate the scalability of these solutions in varied cultural and socioeconomic contexts and their potential to support holistic child development beyond academic settings.

In conclusion, AI-powered robots represent a significant advancement in educational technology, offering tools that enhance learning and promote emotional and social growth. This study provides a foundation for integrating these technologies into early education, reaffirming their potential to create more inclusive, engaging, and impactful learning environments.

\balance
\bibliographystyle{ACM-Reference-Format}
\bibliography{IDC_Preprint}

\end{document}